\documentclass[11pt,a4paper]{article}
\usepackage[hyperref]{acl2020}
\usepackage{times}
\usepackage{latexsym}
\usepackage{graphicx}
\usepackage{multirow}
\usepackage{url}
\usepackage{amsmath}
\usepackage{todonotes}
\usepackage{booktabs}
\usepackage{comment}
\usepackage{enumitem}
\usepackage{tabu}
\usepackage{dsfont}

\usepackage{microtype}

\usepackage{algorithm}
\usepackage{algorithmic}

\renewcommand{\algorithmiccomment}[1]{\bgroup\hfill\small//~#1\egroup}

\DeclareMathOperator*{\argmax}{argmax}

\usepackage{url}
\aclfinalcopy 
\def\aclpaperid{561} 

\title{Neural Syntactic Preordering for Controlled Paraphrase Generation}

\author{Tanya Goyal \and Greg Durrett \\
  Department of Computer Science \\
  The University of Texas at Austin \\
  {\tt tanyagoyal@utexas.edu, gdurrett@cs.utexas.edu}}

\date{}

\begin{document}
\maketitle
\begin{abstract}
Paraphrasing natural language sentences 
is a multifaceted process: it might involve replacing individual words or short phrases, local rearrangement of content, or high-level restructuring like topicalization or passivization. Past approaches struggle to cover this space of paraphrase possibilities in an interpretable manner. Our work, inspired by pre-ordering literature in machine translation, uses syntactic transformations to softly ``reorder'' the source sentence and guide our neural paraphrasing model. First, given an input sentence, we derive a set of feasible syntactic rearrangements using an encoder-decoder model. This model operates over a partially lexical, partially syntactic view of the sentence and can reorder big chunks. Next, we use each proposed rearrangement to produce a sequence of position embeddings, which encourages our final encoder-decoder paraphrase model to attend to the source words in a particular order. Our evaluation, both automatic and human, shows that the proposed system retains the quality of the baseline approaches while giving a substantial increase in the diversity of the generated paraphrases.\footnote{Data and code are available at \url{https://github.com/tagoyal/sow-reap-paraphrasing}}
\end{abstract}

\section{Introduction}
Paraphrase generation \cite{mckeown1983paraphrasing, barzilay2003learning} has seen a recent surge of interest, both with large-scale dataset collection and curation \cite{lan-etal-2017-continuously,wieting-gimpel-2018-paranmt} and with modeling advances such as deep generative models \cite{gupta2018deep, li2019decomposable}. Paraphrasing models have proven to be especially useful if they expose control mechanisms that can be manipulated to produce diverse paraphrases \cite{iyyer2018adversarial,chen-etal-2019-controllable, park2019paraphrase}, which allows these models to be employed for data augmentation \cite{wei2018fast} and adversarial example generation \cite{iyyer2018adversarial}. However, prior methods involving syntactic control mechanisms do not effectively cover the space of paraphrase possibilities. Using syntactic templates covering the top of the parse tree \cite{iyyer2018adversarial} is inflexible, and using fully-specified exemplar sentences \cite{chen-etal-2019-controllable} poses the problem of how to effectively retrieve such sentences. For a particular input sentence, it is challenging to use these past approaches to enumerate the set of reorderings that make sense \emph{for that sentence.}

\begin{figure}
\centering
    \includegraphics[trim=172mm 190mm 135mm 90mm,scale=0.24,clip]{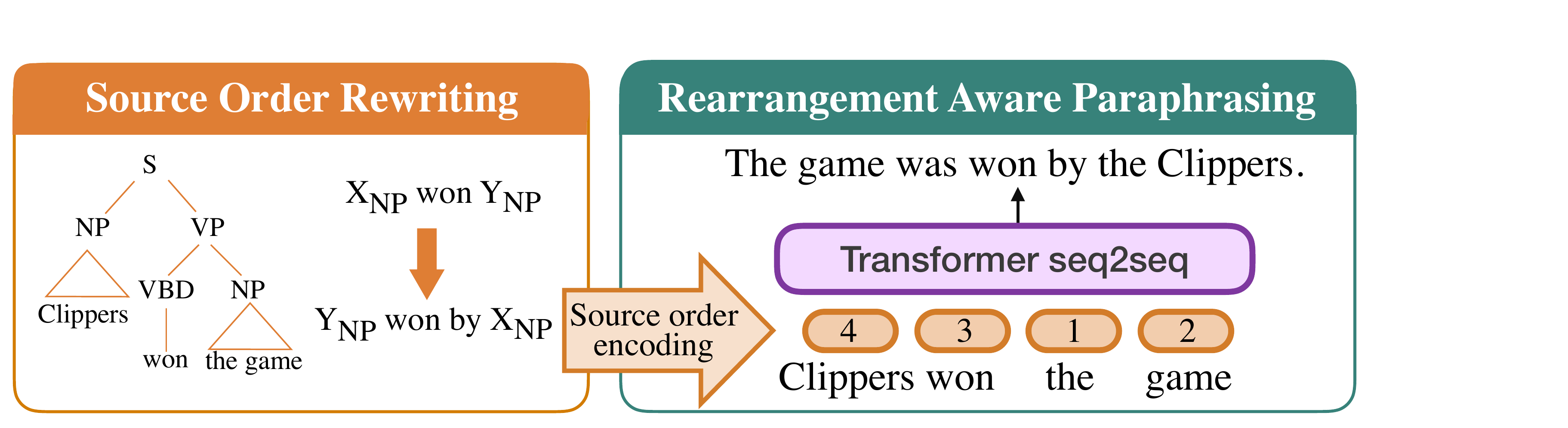}
    \caption{Overview of our paraphrase model. First, we choose various pairs of constituents to abstract away in the source sentence, then use a neural transducer to generate possible reorderings of the abstracted sentences. From these, we construct a guide reordering of the input sentence which then informs the generation of output paraphrases.}
    \label{fig:system-overview}
\end{figure} 

\begin{figure}
\centering
    \includegraphics[trim=40mm 145mm 0mm 40mm, scale=0.24, clip]{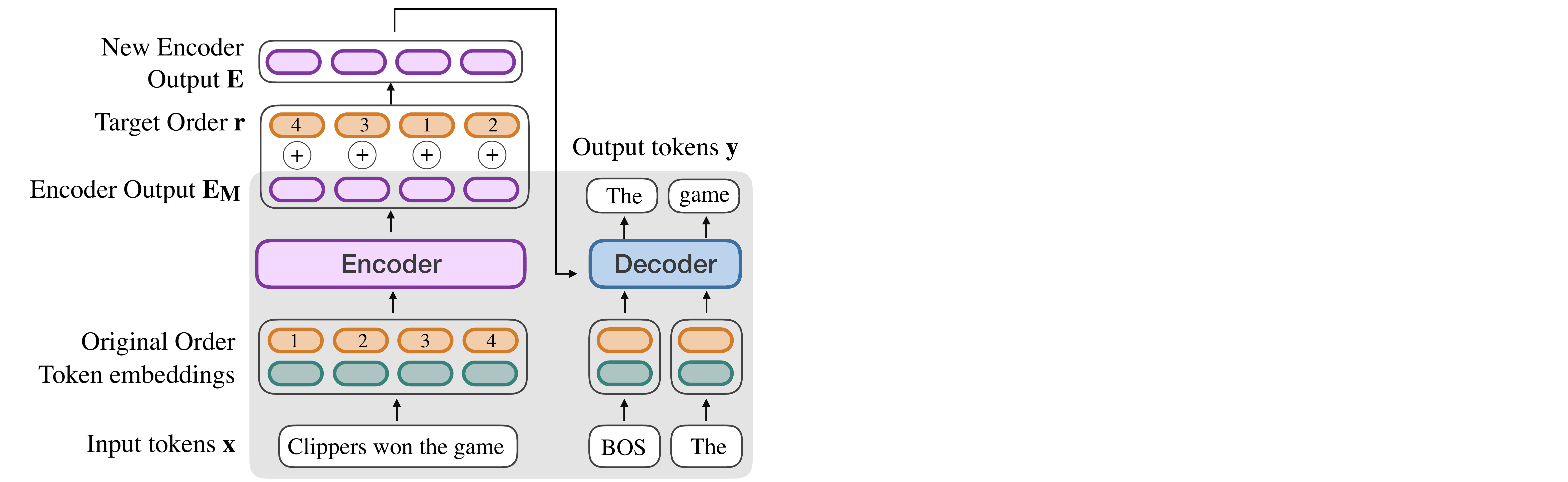}
    \caption{Rearrangement aware paraphrasing (\textsc{Reap}) model. The gray area corresponds to the standard transformer encoder-decoder system. Our model adds position embeddings corresponding to the target reordering to encoder outputs. The decoder attends over these augmented encodings during both training and inference.} 
    \label{fig:position-model}
\end{figure}

In this paper, we propose a two-stage approach to address these limitations, outlined in Figure \ref{fig:system-overview}. First, we use an encoder-decoder model (\textsc{Sow}, for Source Order reWriting) to apply transduction operations over various abstracted versions of the input sentence. These transductions yield possible reorderings of the words and constituents, which can be combined to obtain multiple feasible rearrangements of the input sentence. Each rearrangement specifies an order that we should visit words of the source sentence; note that such orderings could encourage a model to passivize (visit the object before the subject), topicalize, or reorder clauses. These orderings are encoded for our encoder-decoder paraphrase model (\textsc{Reap}, for REarrangement Aware Paraphrasing) by way of position embeddings, which are added to the source sentence encoding to specify the desired order of generation (see Figure~\ref{fig:position-model}). This overall workflow is inspired by the pre-ordering literature in machine translation \cite{xia2004improving,collins2005clause}; however, our setting explicitly requires entertaining a \emph{diverse} set of possible orderings corresponding to different paraphrasing phenomena. 

We train and evaluate our approach on the large-scale English paraphrase dataset \textsc{ParaNMT-50M} \cite{wieting-gimpel-2018-paranmt}. Results show that our approach generates considerably more diverse paraphrases while retaining the quality exhibited by strong baseline models. We further demonstrate that the proposed syntax-based transduction procedure generates a feasible set of rearrangements for the input sentence. Finally, we show that position embeddings provide a simple yet effective way to encode reordering information, and that the generated paraphrases exhibit high compliance with the desired reordering input. 

\section{Method}
Given an input sentence $\mathbf{x}=\{x_1,x_2,\dots,x_n\}$, our goal is to generate a set of structurally distinct paraphrases $Y=\{\mathbf{y}^1, \mathbf{y}^2,\dots,\mathbf{y}^k\}$. We achieve this by first producing $k$ diverse reorderings for the input sentence, $R=\{\mathbf{r}^1,\mathbf{r}^2,\dots,\mathbf{r}^k\}$, that guide the generation order of each corresponding $\mathbf{y}$. Each reordering is represented as a permutation of the source sentence indices.

Our method centers around a sequence-to-sequence model which can generate a paraphrase roughly respecting a particular ordering of the input tokens. Formally, this is a model $P(\mathbf{y} \mid \mathbf{x},\mathbf{r})$. First, we assume access to the set of target reorderings $R$ and describe this rearrangement aware paraphrasing model (\textsc{Reap}) in Section \ref{sec:reap-model}. Then, in Section \ref{sec:rearrangement-model}, we outline our reordering approach, including the source order rewriting (\textsc{Sow}) model, which produces the set of reorderings appropriate for a given input sentence $\mathbf{x}$ during inference $(\mathbf{x} \rightarrow R)$.

\begin{figure*}
\centering
    \includegraphics[trim=10mm 150mm 0mm 5mm, scale=0.24,clip]{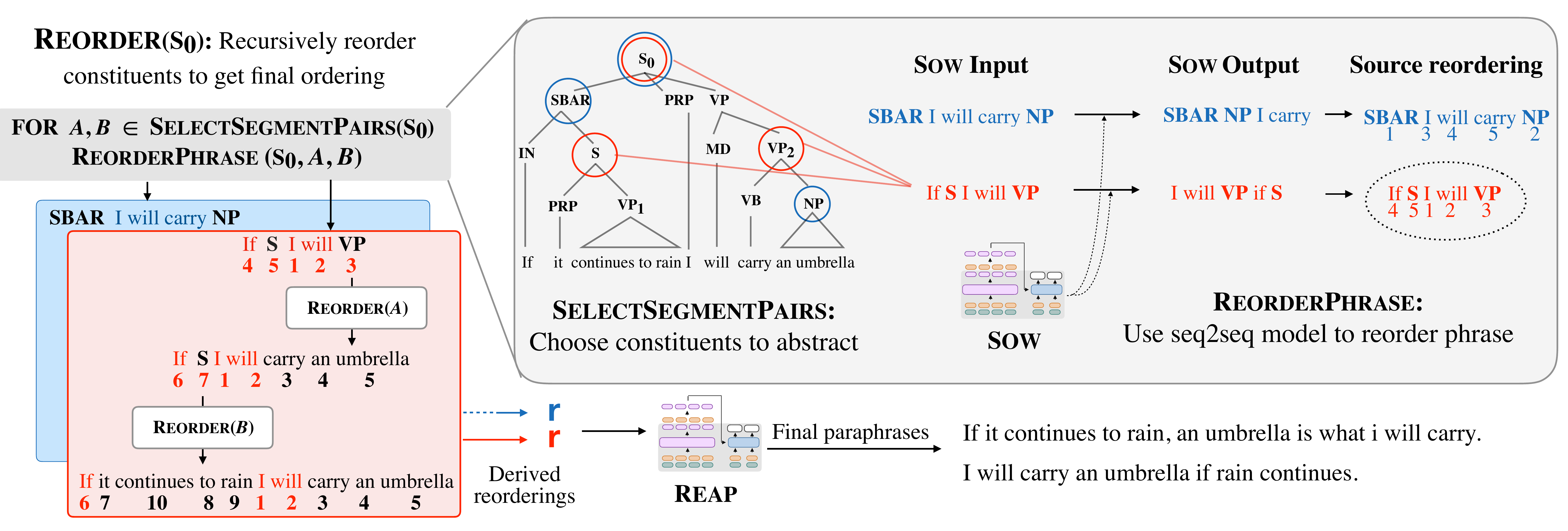}
    \caption{Overview of the source sentence rearrangement workflow for one level of recursion at the root node. First, candidate tree segment pairs contained within the input node are selected. A transduction operation is applied over the abstracted phrase, giving the reordering \textbf{4 5 1 2 3} for the case shown in red, then the process recursively continues for each abstracted node. This results in a reordering for the full source sentence; the reordering indices serve as additional input to the \textsc{Reap} model.}
    \label{fig:rearrangement-workflow}
\end{figure*}

\subsection{Base Model}
\label{sec:base-model}
The models discussed in this work build on a standard sequence-to-sequence transformer model \cite{vaswani2017attention} that uses stacked layers of self-attention to both encode the input tokens $\mathbf{x}$ and decode the corresponding target sequence $\mathbf{y}$. This model is pictured in the gray block of Figure~\ref{fig:position-model}. Throughout this work, we use byte pair encoding (BPE) \cite{sennrich2016neural} to tokenize our input and output sentences. These models are trained in the standard way, maximizing the log likelihood of the target sequence using teacher forcing.  Additionally, in order to ensure that the decoder does not attend to the same input tokens repeatedly at each step of the decoding process, we include a coverage loss term, as proposed in \citet{see2017get}.  

Note that since the architecture of the transformer model is non-recurrent, it adds position embeddings to the input word embeddings in order to indicate the correct sequence of the words in both $\mathbf{x}$ and $\mathbf{y}$ (see Figure \ref{fig:position-model}). In this work, we propose using an additional set of position embeddings to indicate the desired order of words during generation, described next.

\subsection{Rearrangement aware Paraphrasing Model (\textsc{Reap})} 
\label{sec:reap-model}
Let $\mathbf{r}=\{r_1,r_2,\dots,r_n\}$ indicate the target reordering corresponding to the input tokens $\mathbf{x}$. We want the model to approximately attend to tokens in this specified order when generating the final output paraphrase. For instance, in the example in Figure \ref{fig:system-overview}, the reordering specifies that when producing the paraphrase, the model should generate content related to \textit{the game} before content related to \textit{Clippers} in the output. In this case, based on the rearrangement being applied, the model will most likely use passivization in its generation, although this is not strictly enforced.

The architecture for our model $P(\mathbf{y} \mid \mathbf{x},\mathbf{r})$ is outlined in Figure \ref{fig:position-model}. Consider an encoder-decoder architecture with a stack of $M$ layers in the encoder and $N$ layers in the decoder. We make the target reordering $\mathbf{r}$ accessible to this transformer model through an additional set of positional embeddings $PE_\mathbf{r}$. We use the sinusoidal function to construct these following \citet{vaswani2017attention}.

Let $E_M=\textrm{encoder}_M(\mathbf{x})$ be the output of the $M^{th}$ (last) layer of the encoder. The special-purpose position embeddings are added to the output of this layer (see Figure \ref{fig:position-model}): $E = E_M + PE_r$. Note that these are separate from standard position embeddings added at the input layer; such embeddings are also used in our model to encode the original order of the source sentence. The transformer decoder model attends over $E$ while computing attention and the presence of the position embeddings should encourage the generation to obey the desired ordering $\mathbf{r}$, while still conforming to the decoder language model. Our experiments in Section \ref{sec:ablations}
show that this position embedding method is able to successfully guide the generation of paraphrases, conditioning on both the input sentence semantics as well as the desired ordering.

\subsection{Sentence Reordering}
\label{sec:rearrangement-model}
We now outline our approach for generating these desired reorderings $\mathbf{r}$. We do this by predicting phrasal rearrangements with the \textsc{Sow} model at various levels of syntactic abstraction of the sentence. We combine multiple such  phrase-level rearrangements to obtain a set $R$ of sentence-level rearrangements. This is done using a top-down approach, starting at the root node of the parse tree. The overall recursive procedure is outlined in Algorithm~\ref{algo:combining-phrase-rearrangements}.

\begin{algorithm}
\caption{\textsc{Reorder}$(t)$}
\begin{algorithmic} 
\REQUIRE Sub-tree $t$ of the input parse tree
\ENSURE Top-$k$ list of reorderings for $t$'s yield
\STATE $\mathcal{T} =$ \textsc{SelectSegmentPairs}$(t)$ \COMMENT{Step 1}
\STATE $\mathcal{R} = $ \textsc{InitializeBeam}$(\text{size} = k)$
\FOR{ $(A, B)$ in $\mathcal{T}$ }
     \STATE $z = $ \textsc{ReorderPhrase}$(t, A, B)$ \COMMENT{Step 2}
    \STATE $R_A(1,\ldots,k) = $  \textsc{Reorder}$(t_A)$ \COMMENT{$k$ orderings}
    \STATE $R_B(1,\ldots,k) = $  \textsc{Reorder}$(t_B)$ \COMMENT{$k$ orderings}
    \FOR{ $r_a,r_b$ in $R_A \times R_B$ }
            \STATE $r = $ \textsc{Combine}$(z, r_a, r_b)$ \COMMENT{Step 3}
            \STATE $\textrm{score}(r)\hspace{-0.1mm}=\hspace{-0.1mm}\textrm{score}(z)+ \textrm{score}(r_a)+\textrm{score}(r_b)$
            \vspace{-4mm}
            \STATE $\mathcal{R}.$push$(r, \textrm{score}(r))$
    \ENDFOR
\ENDFOR
\RETURN $\mathcal{R}$
\end{algorithmic}
\label{algo:combining-phrase-rearrangements}
\end{algorithm}

One step of the recursive algorithm has three major steps: Figure \ref{fig:rearrangement-workflow} shows the overall workflow for one iteration (here, the root node of the sentence is selected for illustration). 
First, we select sub-phrase pairs of the input phrase that respect parse-tree boundaries, where each pair consists of non-overlapping phrases (Step 1). Since the aim is to learn generic syntax-governed rearrangements, we abstract out the two sub-phrases, and replace them with non-terminal symbols, retaining only the constituent tag information. For example, we show three phrase pairs in Figure~\ref{fig:rearrangement-workflow} that can be abstracted away to yield the reduced forms of  the sentences. We then use a seq2seq model to obtain rearrangements for each abstracted phrase (Step 2). Finally, this top-level rearrangement is combined with recursively-constructed phrase rearrangements within the abstracted phrases to obtain sentence-level rearrangements (Step 3).

\subsubsection*{Step 1: \textsc{SelectSegmentPairs}}
We begin by selecting phrase tuples that form the input to our seq2seq model. A phrase tuple $(t, A, B)$ consists of a sub-tree $t$ with the constituents $A$ and $B$ abstracted out (replaced by their syntactic categories). 
For instance, in Figure \ref{fig:rearrangement-workflow}, the S$_0$, S, and VP$_2$ nodes circled in red form a phrase tuple. Multiple distinct combinations of $A$ and $B$ are possible.\footnote{In order to limit the number of such pairs, we employ a threshold on the fraction of non-abstracted words remaining in the phrase, outlined in more detail in the Appendix.}  

\subsubsection*{Step 2: \textsc{ReorderPhrase}} 
Next, we obtain rearrangements for each phrase tuple $(t, A, B)$. We first form an input consisting of the yield of $t$ with $A$ and $B$ abstracted out; e.g. \textit{If S I will VP}, shown in red in Figure~\ref{fig:rearrangement-workflow}. We use a sequence-to-sequence model (the \textsc{Sow} model) that takes this string as input and produces a corresponding output sequence. We then perform word-level alignment between the input and generated output sequences (using cosine similarity between GloVe embeddings) to obtain the rearrangement that must be applied to the input sequence.\footnote{We experimented with a pointer network to predict indices directly; however, the approach of generate and then align post hoc resulted in a much more stable model.} The log probability of the output sequence serves as a score for this rearrangement.

\paragraph{\textsc{Sow} model} The \textsc{Sow} model is a sequence-to-sequence model $P(\mathbf{y'} \mid \mathbf{x'},o)$, following the transformer framework in Section~\ref{sec:base-model}.\footnote{See Appendix for \textsc{Sow} model architecture diagram.} Both $\mathbf{x'}$ and $\mathbf{y'}$ are encoded using the word pieces vocabulary; additionally, embeddings corresponding to the POS tags and constituent labels (for non-terminals) are added to the input embeddings. 
 
For instance, in Figure \ref{fig:rearrangement-workflow}, \textit{If S I will VP} and \textit{I will VP if S} is an example of an $(\mathbf{x'}, \mathbf{y'})$, pair. While not formally required, Algorithm~\ref{algo:combining-phrase-rearrangements} ensures that there are always exactly two non-terminal labels in these sequences. $o$ is a variable that takes values \textsc{Monotone} or \textsc{Flip}. This encodes a preference to keep the two abstracted nodes in the same order or to ``flip'' them in the output.\footnote{In syntactic translation systems, rules similarly can be divided by whether they preserve order or invert it \cite{wu1997stochastic}.} $o$ is encoded in the model with additional positional encodings of the form $\{\dots0,0,1,0,\dots2,0\dots\}$ for monotone and $\{\dots0,0,2,0,\dots1,0\dots\}$ for flipped, wherein the non-zero positions correspond to the positions of the abstracted non-terminals in the phrase. These positional embeddings for the \textsc{Sow Model} are handled analogously to the $\mathbf{r}$ embeddings for the \textsc{Reap} model. During inference, we use both the monotone rearrangement and flip rearrangement to generate two reorderings, one of each type, for each phrase tuple.

We describe training of this model in Section \ref{sec:dataset}.

\subsubsection*{Step 3: \textsc{Combine}}
The previous step gives a rearrangement for the subtree $t$. To obtain a sentence-level rearrangement from this, we first recursively apply the \textsc{Reorder} algorithm on subtrees $t_A$ and $t_B$ which returns the top-k rearrangements of each subtree. We iterate over each rearrangement pair $(r_a,r_b)$, applying these reorderings to the abstracted phrases $A$ and $B$. This is illustrated on the left side of Figure \ref{fig:rearrangement-workflow}. The sentence-level representations, thus obtained, are scored by taking a mean over all the phrase-level rearrangements involved.

\section{Data and Training}
\label{sec:dataset}

We train and evaluate our model on the \textsc{ParaNMT-50M} paraphrase dataset \cite{wieting-gimpel-2018-paranmt} constructed by backtranslating the Czech sentences of the CzEng \cite{bojar2016czeng} corpus. We filter this dataset to remove shorter sentences (less than 8 tokens), low quality paraphrase pairs (quantified by a translation score included with the dataset) and examples that exhibit low reordering (quantified by a reordering score based on the position of each word in the source and its aligned word in the target sentence). This leaves us with over 350k paired paraphrase pairs.

\subsection{Training Data for \textsc{Reap}}
To train our \textsc{Reap} model (outlined in Section \ref{sec:reap-model}), we take existing paraphrase pairs $(\mathbf{x},\mathbf{y}^*)$ and derive pseudo-ground truth rearrangements $\mathbf{r}^*$ of the source sentence tokens based on their alignment with the target sentence. To obtain these rearrangements, we first get contextual embeddings \cite{devlin2019bert} for all tokens in the source and target sentences. We follow the strategy outlined in \citet{lerner2013source} and perform reorderings as we traverse down the dependency tree. Starting at the root node of the source sentence, we determine the order between the head and its children (independent of other decisions) based on the order of the corresponding aligned words in the target sentence. We continue this traversal recursively to get the sentence level-rearrangement. This mirrors the rearrangement strategy from Section \ref{sec:rearrangement-model}, which operates over constituency parse tree instead of the dependency parse.

Given triples $(\mathbf{x},\mathbf{r}^*,\mathbf{y}^*)$, we can train our \textsc{Reap} model to generate the final paraphrases conditioning on the pseudo-ground truth reorderings.

\subsection{Training Data for \textsc{Sow}} 

\begin{figure}
\centering
    \includegraphics[trim=175mm 235mm 200mm 75mm,scale=0.23,clip]{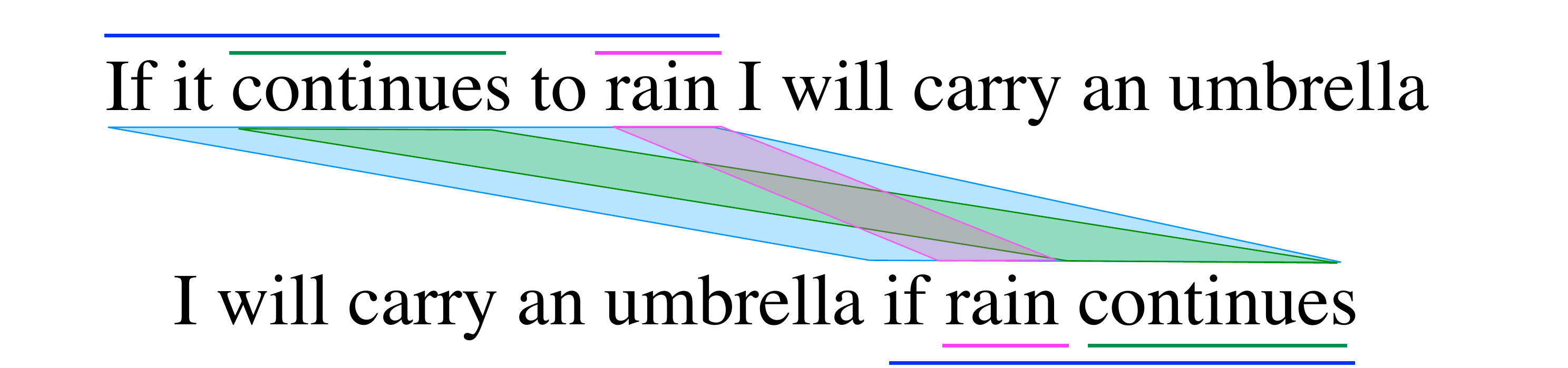}
    \caption{Paraphrase sentence pair and its aligned tuples $A \rightarrow B,C$ and  $A' \rightarrow B',C'$. These produce the training data for the \textsc{Sow Model}.} 
    \label{fig:sow-training}
\end{figure}

The \textsc{ParaNMT-50M} dataset contains sentence-level paraphrase pairs. However, in order to train our \textsc{Sow} model (outlined in section \ref{sec:rearrangement-model}), we need to see phrase-level paraphrases with syntactic abstractions in them. We extract these from the \textsc{ParaNMT-50M} dataset using the following procedure, shown in Figure~\ref{fig:sow-training}. We follow \citet{zhang2019bertscore} and compute a phrase alignment score between all pairs of constituents in a sentence and its paraphrase.\footnote{The score is computed using a weighted mean of the contextual similarity between individual words in the phrases, where the weights are determined by the corpus-level inverse-document frequency of the words. Details in the Appendix.} From this set of phrase alignment scores, we compute a partial one-to-one mapping between phrases (colored shapes in Figure~\ref{fig:sow-training}); that is, not all phrases get aligned, but the subset that do are aligned one-to-one. Finally, we extract aligned chunks similar to rule alignment in syntactic translation \cite{galley2004translation}: when aligned phrases $A$ and $A'$ subsume aligned phrase pairs $(B,C)$ and $(B', C')$ respectively, we can extract the aligned tuple $(t_A, B, C)$ and $(t_{A'},B',C')$. The phrases  $(B,C)$ and $(B', C')$ are abstracted out to construct training data for the phrase-level transducer, including supervision of whether $o =$ \textsc{Monotone} or \textsc{Flip}.
Using the above alignment strategy, we were able to obtain over $1$ million aligned phrase pairs.

\begin{table*}[]
\centering
\small
\begin{tabular}{r|cccc|c|cc}
\toprule
Model & \multicolumn{4}{c|}{oracle quality (over 10 sentences, no rejection) $\uparrow$} & & \multicolumn{2}{|c}{pairwise diversity (post-rejection)} \\ \midrule
 & BLEU & ROUGE-1 & ROUGE-2 & ROUGE-L & \% rejected & self-BLEU $\downarrow$ & self-WER $\uparrow$  \\ \midrule
copy-input & $18.4$ & $54.4$ & $27.2$ & $49.2$ & $0$ & $-$ & $-$ \\
\textsc{SCPN} & $21.3$ & $53.2$ & $30.3$ & $51.0$ & $40.6$ &  $35.9$ & $63.4$ \\
Transformer seq2seq & $32.8$ & $63.1$ & $41.4$ & $63.3$ & $12.7$ & $50.7$ & $35.4$ \\  
 + diverse-decoding & $24.8$ & $56.8$ & $33.2$ & $56.4$   & $21.3$ & $34.2$ & $58.1$ \\ \midrule
\textsc{Sow-Reap (LSTM)}  & $27.0$ & $57.9$ & $34.8$ & $57.5$ & $31.7$ & $46.2$ & $53.9$\\
\textsc{Sow-Reap} & $30.9$ & $62.3$ & $40.2$ & $61.7$  & $15.9$ & $38.0$ & $57.9$ \\  
\bottomrule
\end{tabular} 
\caption{Quality and diversity metrics for the different models. Our proposed approach outperforms other diverse models (SCPN and diverse decoding) in terms of all the quality metrics. These models exhibit higher diversity, but with many more rejected paraphrases, indicating that these models more freely generate bad paraphrases.}
\label{table:paraphrase-quality}
\end{table*} 

\section{Evaluation} 

\paragraph{Setup} As our main goal is to evaluate our model's ability to generate diverse paraphrases, we obtain a set of paraphrases and compare these to sets of paraphrases produced by other methods. To obtain 10 paraphrases, we first compute a set of 10 distinct reorderings $\mathbf{r}^1,\ldots,\mathbf{r}^{10}$ with the \textsc{Sow} method from Section~\ref{sec:rearrangement-model} and then use the \textsc{Reap} to generate a 1-best paraphrase for each. We use top-$k$ decoding to generate the final set of paraphrases corresponding to the reorderings. Our evaluation is done over 10k examples from \textsc{ParaNMT-50M}.

\subsection{Quantitative Evaluation} 
\paragraph{Baselines} We compare our model against the Syntactically Controlled Paraphrase Network (\textbf{SCPN}) model proposed in prior work \cite{iyyer2018adversarial}. It produces $10$ distinct paraphrase outputs conditioned on a pre-enumerated list of syntactic templates. This approach has been shown to outperform other paraphrase approaches that condition on interpretable intermediate structures \cite{chen-etal-2019-controllable}. Additionally, we report results on the following baseline models: i) A \textbf{copy-input} model that outputs the input sentence exactly. ii) A vanilla \textbf{seq2seq} model that uses the same transformer encoder-decoder architecture from Section \ref{sec:base-model} but does not condition on any target rearrangement. We use top-$k$ sampling \cite{fan2018hierarchical} to generate $10$ paraphrases from this model.\footnote{Prior work \cite{wang2019task,li2019decomposable} has shown that such a transformer-based model provides a strong baseline and outperforms previous LSTM-based \cite{hasan2016neural} and VAE-based \cite{gupta2018deep} approaches.} iii) A \textbf{diverse-decoding} model that uses the above transformer seq2seq model with diverse decoding \cite{kumar2019submodular} during generation. Here, the induced diversity is uncontrolled and aimed at maximizing metrics such as distinct n-grams and edit distance between the generated sentences. 
iv) A \textbf{LSTM} version of our model where the \textsc{Reap} model uses LSTMs with attention \cite{Bahdanau2014NeuralMT} and copy \cite{see2017get} instead of transformers. We still use the transformer-based phrase transducer to obtain the source sentence reorderings, and still use positional encodings in the LSTM attention.

Similar to \citet{cho2019mixture}, we report two types of metrics: 
\setlist{nolistsep}
\begin{enumerate}[leftmargin=*]
\item \textbf{Quality}: Given $k$ generated paraphrases $Y=\{\mathbf{y}^1, \mathbf{y}^2 \dots \mathbf{y}^k \}$ for each input sentence in the test set, we select $\hat{\mathbf{y}}^{best}$ that achieves the best (oracle) sentence-level score with the ground truth paraphrase $\mathbf{y}$. The corpus level evaluation is performed using pairs $(\hat{\mathbf{y}}^{best}, \mathbf{y})$.
\item \textbf{Diversity}: We calculate BLEU or WER between all pairs $(\mathbf{y}^i, \mathbf{y}^j)$ generated by a single model on a single sentence, then macro-average these values at a corpus-level. 
\end{enumerate}
In addition to these metrics, we use the paraphrase similarity model proposed by \citet{wieting2017learning} to compute a paraphrase score for generated outputs with respect to the input. Similar to \citet{iyyer2018adversarial}, we use this score to filter out low quality paraphrases. We report on the rejection rate according to this criterion for all models. Note that our diversity metric is computed \textit{after} filtering as it is easy to get high diversity by including nonsensical paraphrase candidates that differ semantically.  

Table \ref{table:paraphrase-quality} outlines the performance of the different models. The results show that our proposed model substantially outperforms the SCPN model across all quality metrics.\footnote{The difference in performance between our proposed model and baseline models is statistically significant according to a paired bootstrap test.} Furthermore, our LSTM model also beats the performance of the SCPN model, demonstrating that the gain in quality cannot completely be attributed to the use of transformers. The quality of our full model (with rearrangements) is also comparable to the quality of the vanilla seq2seq model (without rearrangements). This demonstrates that the inclusion of rearrangements from the syntax-based neural transducer do not hurt quality, while leading to a substantially improved diversity performance. 

The SCPN model has a high rejection score of $40.6\%$. This demonstrates that out of the $10$ templates used to generate paraphrases for each sentence, on average $4$ were not appropriate for the given sentence, and therefore get rejected. On the other hand, for our model, only $15.9\%$ of the generated paraphrases get rejected, implying that the rearrangements produced were generally meaningful. This is comparable to the $12.7\%$ rejection rate exhibited by the vanilla seq2seq model that does not condition on any syntax or rearrangement, and is therefore never obliged to conform to an inappropriate structure. 

Finally, our model exhibits a much higher diversity within the generated paraphrases compared to the transformer seq2seq baseline. As expected, the SCPN model produces slightly more diverse paraphrases as it explicitly conditions the generations on templates with very different top level structures. However, this is often at the cost of semantic equivalence, as demonstrated by both quantitative and human evaluation (next section).  A similar trend was observed with the diverse-decoding scheme. Although it leads to more diverse generations, there is a substantial decrease in quality compared to \textsc{Sow-Reap} and the seq2seq model. Moreover, the paraphrases have a higher rejection rate (21.3\%), suggesting that diverse decoding is more likely to produce nonsensical paraphrases. A similar phenomenon is also reported by \citet{iyyer2018adversarial}, wherein diverse-decoding resulted in paraphrases with different semantics than the input.

\begin{table*}[h]
\small
\begin{tabular}{p{2.6cm}p{5.7cm}p{6.5cm}}
\toprule
Input & \textsc{Sow-Reap} & \textsc{SCPN} \\ \midrule
\multirow{1}{0.35\columnwidth}{if at any time in the preparation of this product the integrity of this container is compromised it should not be used .} & this container should not be used if any time in the preparation of this product is compromised & in the preparation of this product , the integrity of this container is compromised , but it should not be used . \\
& if the integrity of the packaging is impaired at any time , the product should not be used .  & where is the integrity of this product of this container the integrity of this container should not be used .\\
& if the product integrity of this container is compromised it should not be used .  & i should not use if at any time in the preparation of this product , it should not be used .\\
\midrule
\multirow{1}{0.35\columnwidth}{i was the first grower to use hydroponics .} & to use hydroponics , i was the first one . & where did i have the first tendency to use hydroponics ? \\
& i used hydroponics for the first time . &  i used to use hydroponics .\\
& to use hydroponics the first time i was . & first i was the first grower to use hydroponics \\
\bottomrule
\end{tabular}
\caption{Examples of paraphrases generated by our system and the baseline SCPN model. Our model successfully rearranges the different structural components of the input sentence to obtain meaningful rearrangements. SCPN conforms to pre-enumerated templates that may not align with a given input.} 
\label{table:paraphrase-quality-examples}
\end{table*}

\paragraph{Syntactic Exemplars}
In addition to SCPN, we compare our proposed model against the controllable generation method of \newcite{chen-etal-2019-controllable}. Their model uses an exemplar sentence as a syntactic guide during generation; the generated paraphrase is trained to incorporate the semantics of the input sentence while emulating the syntactic structure of the exemplar (see Appendix~\ref{sec:syntactic_exemplars} for examples). However, their proposed approach depends on the availability of such exemplars at test time; they manually constructed these for their test set ($800$ examples). Since we do not have such example sentences available for our test data, we report results of our model's performance on their test data.

Note that \citet{chen-etal-2019-controllable} carefully curated the exemplar to be syntactically similar to the actual target paraphrase. Therefore, for fair comparison, we report results using the ground truth ordering (that similarly leverages the target sentence to obtain a source reordering), followed by the \textsc{Reap} model. This model (ground truth order + \textsc{Reap}) achieves a 1-best BLEU score of 20.9, outperforming both the prior works: \citet{chen-etal-2019-controllable} (13.6 BLEU) and SCPN (17.8 BLEU with template, 19.2 BLEU with full parse). Furthermore, our full \textsc{Sow-Reap} model gets an oracle-BLEU (across 10 sentences) score of 23.8. These results show that our proposed formulation outperforms other controllable baselines, while being more flexible.

\subsection{Qualitative Evaluation} 
Table \ref{table:paraphrase-quality-examples} provides examples of paraphrase outputs produced by our approach and SCPN. The examples show that our model exhibits syntactic diversity while producing reasonable paraphrases of the input sentence. On the other hand, SCPN tends to generate non-paraphrases in order to conform to a given template, which contributes to increased diversity but at the cost of semantic equivalence. In Table~\ref{table:qual-rule-seq}, we show the corresponding sequence of rules that apply to an input sentence, and the final generated output according to that input rearrangement. Note that for our model, on average, $1.8$ phrase-level reorderings were combined to produce sentence-level reorderings (we restrict to a maximum of $3$). More examples along with the input rule sequence (for our model) and syntactic templates (for SCPN) are provided in the Appendix.

\begin{table}[]
\small
\begin{tabular}{p{0.95\columnwidth}}
\toprule
\textbf{Input Sentence}: if at any time in the preparation of this product the integrity of this container is compromised it should not be used . \\ \midrule
\textbf{Rule Sequence}: 
if S it should not VB used . $\rightarrow$ should not VB used if S  (parse tree level: 0)\\ \vspace{0.001mm}
at NP the integrity of this container VBZ compromised $\rightarrow$ this container VBZ weakened at NP (parse tree level: 1)\\ \vspace{0.001mm}
the NN of NP $\rightarrow$ NP NN (parse tree level: 2)\\
\midrule
\textbf{Generated Sentence}: this container should not be used if the product is compromised at any time in preparation . \\
\bottomrule
\end{tabular} 
\caption{Examples of our model's rearrangements applied to a given input sentence. Parse tree level indicates the rule subtree's depth from the root node of the sentence. The \textsc{Reap} model's final generation considers the rule reordering at the higher levels of the tree but ignores the rearrangement within the lower sub-tree.} 
\label{table:qual-rule-seq}
\end{table}

\paragraph{Human Evaluation} We also performed human evaluation on Amazon Mechanical Turk to evaluate the quality of the generated paraphrases. We randomly sampled $100$ sentences from the development set. For each of these sentences, we obtained $3$ generated paraphrases from each of the following models: i) SCPN, ii) vanilla sequence-to-sequence and iii) our proposed \textsc{Sow-Reap} model. We follow earlier work \cite{kok2010hitting,iyyer2018adversarial} and obtain quality annotations on a 3 point scale: \textbf{0} denotes not a paraphrase, \textbf{1} denotes that the input sentence and the generated sentence are paraphrases, but the generated sentence might contain grammatical errors, \textbf{2} indicates that the input and the candidate are paraphrases. To emulate the human evaluation design in \citet{iyyer2018adversarial}, we sample paraphrases \textit{after} filtering using the criterion outlined in the previous section and obtain three judgements per sentence and its 9 paraphrase candidates. Table \ref{table:human-eval} outlines the results from the human evaluation. As we can see, the results indicate that the quality of the paraphrases generated from our model is substantially better than the SCPN model.\footnote{The difference of our model performance with SCPN is statistically significant, while that with baseline seq2seq is not according to a paired bootstrap test.} Furthermore, similar to quantitative evaluation, the human evaluation also demonstrates that the performance of this model is similar to that of the vanilla sequence-to-sequence model, indicating that the inclusion of target rearrangements do not hurt performance.

\subsection{Ablations and Analysis} 
\label{sec:ablations}
\subsubsection{Evaluation of \textsc{Sow} Model}
Next, we intrinsically evaluate the performance of our \textsc{Sow} model (Section \ref{sec:rearrangement-model}). Specifically, given a budget of 10 reorderings, we want to understand how close our \textsc{Sow} model comes to covering the target ordering. We do this by evaluating the \textsc{Reap} model in terms of oracle perplexity (of the ground truth paraphrase) and oracle BLEU over these 10 orderings.

We evaluate our proposed approach against $3$ systems: a) \textbf{Monotone} reordering $\{1,2,\dots,n\}$. b) \textbf{Random} permutation, by randomly permuting the children of each node as we traverse down the constituency parse tree. c) \textbf{Ground Truth}, using the pseudo-ground truth rearrangement (outlined in Section \ref{sec:dataset}) between the source and ground-truth target sentence. This serves as an upper bound for the reorderings' performance, as obtained by the recursive phrase-level transducer.

\begin{table}[]
\small
\centering
\begin{tabular}{c|ccc}
\toprule
Model & \textbf{2} & \textbf{1} & \textbf{0} \\ \midrule
SCPN \cite{iyyer2018adversarial}  & 35.9 & 24.8 & 39.3 \\
Transformer seq2seq & 45.1 & 20.6 & 34.3 \\ 
\textsc{Sow-Reap} & 44.5 & 22.6 & 32.9 \\
\bottomrule
\end{tabular}
\caption{Human annotated quality across different models. The evaluation was done on a 3 point quality scale, 2 = grammatical paraphrase, 1 = ungrammatical paraphrase, 0 = not a paraphrase.} 
\label{table:human-eval} 
\end{table}

\begin{table}[]
\centering
\small
\begin{tabular}{c|c|c}
\toprule
Ordering & oracle-ppl $\downarrow$ & oracle-BLEU $\uparrow$ \\ \midrule
Monotone & $10.59$ & $27.98$ \\
Random & $9.32$ & $27.10$\\  
\textsc{Sow} & $8.14$ & $30.02$ \\  \midrule
Ground Truth & $7.79$ & $36.40$ \\
\bottomrule
\end{tabular}
\caption{Comparison of different source reordering strategies. Our proposed approach outperforms baseline monotone and random rearrangement strategies.} 
\label{table:phrase-reordering-eval}
\end{table}

Table \ref{table:phrase-reordering-eval} outlines the results for 10 generated paraphrases from each rearrangement strategy. Our proposed approach outperforms the baseline monotone and random reordering strategies. Furthermore, the \textsc{Sow} model's oracle perplexity is close to that of the ground truth reordering's perplexity, showing that the proposed approach is capable of generating a diverse set of rearrangements such that one of them often comes close to the target rearrangement. The comparatively high performance of the ground truth reorderings demonstrates that the positional embeddings are effective at guiding the \textsc{Reap} model's generation.

\begin{figure}
\centering
    \includegraphics[trim=5mm 0mm 0mm 0mm,scale=0.4,clip]{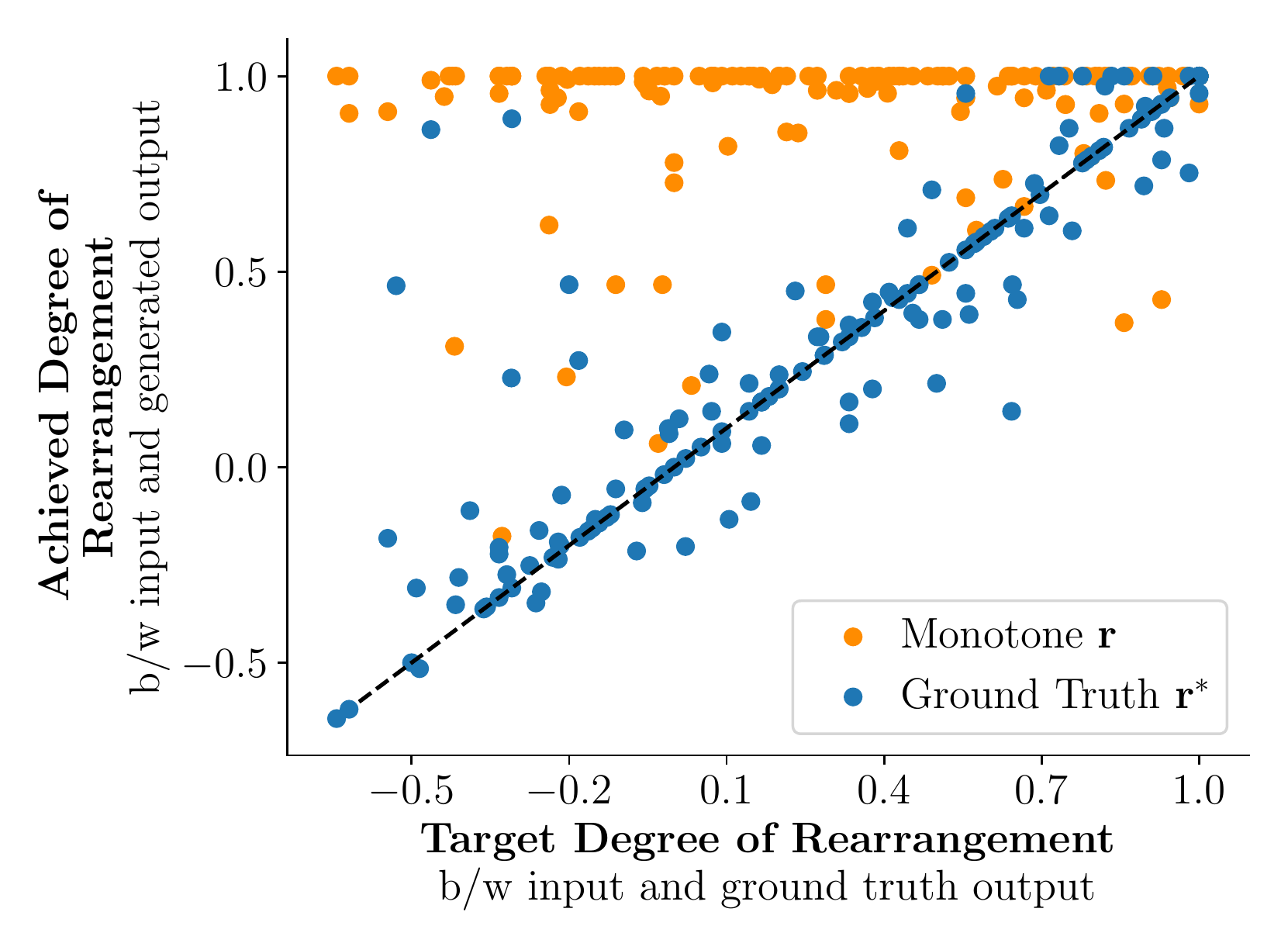}
    \caption{The degree of rearrangement (Kendall's Tau) achieved by conditioning on monotone and  pseudo-ground truth reorderings ($\mathbf{r^*}$). The dotted line denotes the ideal performance (in terms of reordering-compliance) of the \textsc{Reap} model, when supplied with perfect reordering $\mathbf{r^*}$. The actual performance of the \textsc{Reap} model mirrors the ideal performance.} 
    \label{fig:compliance-model}
\end{figure} 
 
\subsubsection{Compliance with target reorderings}
Finally, we evaluate whether the generated paraphrases follow the target reordering \textbf{r}. Note that we do not expect or want our \textsc{Reap} model to be absolutely compliant with this input reordering since the model should be able to correct for the mistakes make by the \textsc{Sow} model and still generate valid paraphrases. Therefore, we perform reordering compliance experiments on only the monotone reordering and the pseudo-ground truth reorderings ($\mathbf{r^*}$, construction outlined in Section \ref{sec:dataset}), since these certainly correspond to valid paraphrases.

For sentences in the test set, we generate paraphrases using monotone reordering and pseudo-ground truth reordering as inputs to \textsc{Reap}. We get the 1-best paraphrase and compute the degree of rearrangement\footnote{Quantified by Kendall's Tau rank correlation between original source order and targeted/generated order. Higher Kendall's Tau indicates lower rearrangement and vice-versa.} between the input sentence and the generated sentence. In Figure \ref{fig:compliance-model}, we plot this as a function of the target degree of rearrangement, i.e., the rearrangement between the input sentence $\mathbf{x}$ and the ground truth sentence $\mathbf{y^*}$. The dotted line denotes the ideal performance of the model in terms of agreement with the perfect reordering $\mathbf{r^*}$. The plot shows that the \textsc{Reap} model performs as desired; the monotone generation results in high Kendall's Tau between input and output. Conditioning on the pseudo-ground truth reorderings ($\mathbf{r^*}$) produces rearrangements that exhibit the same amount of reordering as the ideal rearrangement. 

\section{Related Work} 
\paragraph{Paraphrase Generation}
Compared to prior seq2seq approaches for paraphrasing \cite{hasan2016neural,gupta2018deep,li-etal-2018-paraphrase}, our model is able to achieve much stronger controllability with an interpretable control mechanism. Like these approaches, we can leverage a wide variety of resources to train on, including backtranslation \cite{pavlick-etal-2015-ppdb, wieting-gimpel-2018-paranmt, parabank2018} or other curated data sources \cite{fader2013paraphrase,lan-etal-2017-continuously}.

\paragraph{Controlled Generation} Recent work on controlled generation aims at controlling attributes such as sentiment \cite{shen2017style}, gender or political slant \cite{prabhumoye2018style}, topic \cite{wang2017steering}, etc. However, these methods cannot achieve fine-grained control over a property like syntax. Prior work on diverse paraphrase generation can be divided into three groups: diverse decoding, latent variable modeling, and syntax-based. The first group uses heuristics such as Hamming distance or distinct $n$-grams to preserve diverse options during beam search decoding \cite{vijayakumar2018diverse,kumar2019submodular}. The second group includes approaches that use uninterpretable latent variables to separate syntax and semantics \cite{chen2019multi}, perturb latent representations to enforce diversity \cite{gupta2018deep,park2019paraphrase} or condition on latent codes used to represent different re-writing patterns \cite{dpage,an2019towards}. \newcite{qian2019exploring} uses distinct generators to output diverse paraphrases. These methods achieve some diversity, but do not control generation in an interpretable manner. Finally, methods that use explicit syntactic structures \cite{iyyer2018adversarial, chen-etal-2019-controllable} may try to force a sentence to conform to unsuitable syntax. Phrase-level approaches \cite{li2019decomposable} are inherently less flexible than our approach.

\paragraph{Machine Translation} Our work is inspired by pre-ordering literature in machine translation. These systems either use hand-crafted rules designed for specific languages \cite{collins2005clause,wang2007chinese} or automatically learn rewriting patterns based on syntax \cite{xia2004improving,dyer2010context, genzel2010automatically,khalilov2011context, lerner2013source}. There also exist approaches that do not rely on syntactic parsers, but induce hierarchical representations to leverage for pre-ordering \cite{tromble2009learning,denero2011inducing}. In the context of translation, there is often a canonical reordering that should be applied to align better with the target language; for instance, head-final languages like Japanese exhibit highly regular syntax-governed reorderings compared to English. However, in diverse paraphrase generation, there doesn't exist a single canonical reordering, making our problem quite different. 

In concurrent work, \citet{chen2020explicit} similarly use an additional set of position embeddings to guide the order of generated words for machine translation. This demonstrates  that the \textsc{Reap} technique is effective for other tasks also. However, they do not tackle the problem of generating plausible reorderings and therefore their technique is less flexible than our full \textsc{Sow-Reap} model. 

\section{Conclusion} 
In this work, we propose a two-step framework for paraphrase generation: construction of diverse syntactic guides in the form of target reorderings followed by actual paraphrase generation that respects these reorderings. Our experiments show that this approach can be used to produce paraphrases that achieve a better quality-diversity trade-off compared to previous methods and strong baselines.

\section*{Acknowledgments}

This work was partially supported by NSF Grant IIS-1814522, a gift from Arm, and an equipment grant from NVIDIA. The authors acknowledge the Texas Advanced Computing Center (TACC) at The University of Texas at Austin for providing HPC resources used to conduct this research. Thanks as well to the anonymous reviewers for their helpful comments.

\bibliography{acl2020}
\bibliographystyle{acl_natbib}

\appendix
\section*{APPENDIX}

\section{\textsc{SelectSegmentPairs}: Limiting number of segment pairs}
As outlined in Section 2.3, the \textsc{SelectSegmentPairs} subroutine returns a set of non-overlapping sub-phrases $(A,B)$. In order to limit the number of sub-phrase pairs during inference, we employ the following heuristics: 

\begin{enumerate}
    \item We compute a score based on number of non-abstracted tokens divided by the total number of tokens in the yield of the parent sub-phrase $t$. We reject pairs $(A,B)$ that have a score of more than $0.6$. This reduces spurious ambiguity by encouraging the model to rearrange big constituents hierarchically rather than only abstracting out small pieces.
    \item We maintain a list of tags that are never individually selected as sub-phrases. These include constituents that would be trivial to the reordering such as determiners (DT), prepositions (IN), cardinal numbers (CD), modals (MD), etc. However, these may be a part of larger constituents that form $A$ or $B$.
\end{enumerate}

\section{Training Data for \textsc{Sow Model}}
In Section 3.2, we outlined our approach for obtaining phrase-level alignments from the \textsc{ParaNMT-50m} dataset used to train the \textsc{Sow Model}. In the described approach, an alignment score is computed between each pair of phrases $p,\hat{p}$ belonging to sentences $s$ and $\hat{s}$ respectively. We use the exact procedure in \citet{zhang2019addressing} to compute the alignment score, outlined below:

\begin{enumerate}
    \item First, we compute an inverse document frequency ($idf$) score for each token in the training set. Let $M = \{s^{(i)}\}$ be the total number of sentences. Then $idf$ of a word $w$ is computed as: 
    \begin{equation*}
        idf(w) = - \log{ \frac{1}{M} \sum_{i=i}^{M}{\mathds{1}[w \in s^{(i)}]}}
    \end{equation*} 
    \item Next, we extract a contextual representation of each word in the two phrases $s$ and $\hat{s}$. We use ELMo \cite{peters2018deep} in our approach.
    \item In order to compute a similarity score between each pair of phrases $(p, \hat{p})$, we use greedy matching to first align each token in the source phrase to its most similar word in the target phrase. To compute phrase-level similarity, these these word-level similarity scores are combined by taking a weighted mean, with weights specified by to the idf scores. Formally, 
    \begin{align*}
        & R_{p,\hat{p}} = \frac{\sum_{w_i \in p}{idf(w_i)\max_{\hat{w_j} \in \hat{p}} w_i^{T}\hat{w_j}}}{\sum_{w_i \in p}{idf(w_i)}} \\
        & P_{p,\hat{p}} = \frac{\sum_{\hat{w_j} \in \hat{p}}{idf(\hat{w_j})\max_{w_i \in p} w_i^{T}\hat{w_j}}}{\sum_{\hat{w_j} \in \hat{p}}{idf(\hat{w_j)}}} \\
        & F_{p,\hat{p}} = \frac{2 P_{p,\hat{p}} R_{p,\hat{p}}}{P_{p,\hat{p}}+R_{p,\hat{p}}}
    \end{align*}
    This scoring procedure is exactly same as the one proposed by \citet{zhang2019bertscore} to evaluate sentence and phrase similarities.
    \item Finally, the phrases $p \in s$ and $\hat{p} \in \hat{s}$ are aligned if: 
    \begin{equation*} 
        p = \argmax_{p_i \in s} F_{p_i,\hat{p}} 
        \;\:  \textrm{\&} 
        \;\:  \hat{p} = \argmax_{\hat{p_j} \in \hat{s}} F_{p,\hat{p_j}}
    \end{equation*} 
\end{enumerate}
These aligned set of phrase pairs $(p,\hat{p})$ are used to construct tuples $(t_A, B, C)$ and $(t_A',B',C')$, as outlined in Section 3.2. Table \ref{table:phrase-examples} provides examples of such phrase pairs.

\begin{table}[t]
\small
\centering
\begin{tabular}{l|l}
\toprule
\textsc{Sow} Input &  \textsc{Sow} Output\\ \midrule
removing the NN from NP & excluding this NN from NP  \\ \\
\begin{tabular}{@{}l@{}}they might consider VP  \\ if NP were imposed\end{tabular}
  & \begin{tabular}{@{}l@{}l@{}}in the case of imposition of  \\  NP , they would consider  \\  VP \end{tabular}
    \\ \\
    \begin{tabular}{@{}l@{}}NP lingered in the  \\ deserted NNS .\end{tabular} &
    \begin{tabular}{@{}l@{}}in the abandoned NNS , \\ there was NP .\end{tabular}
  \\ \\
  \begin{tabular}{@{}l@{}}PP was a black NN \\ archway .\end{tabular}
  &
  \begin{tabular}{@{}l@{}}was a black NN passage  \\  PP .\end{tabular} \\ \\
\begin{tabular}{@{}l@{}}there is already a ring  \\ NN PP .\end{tabular}
  &
PP circular NN exist . \\
\bottomrule
\end{tabular}
\caption{Examples of aligned phrase pairs with exactly two sub-phrases abstracted out and replaced with constituent labels. These phrase pairs are used to train the \textsc{Sow Model}.}
\label{table:phrase-examples}
\end{table}

\begin{figure}
\centering
    \includegraphics[trim=40mm 125mm 0mm 40mm, scale=0.24, clip]{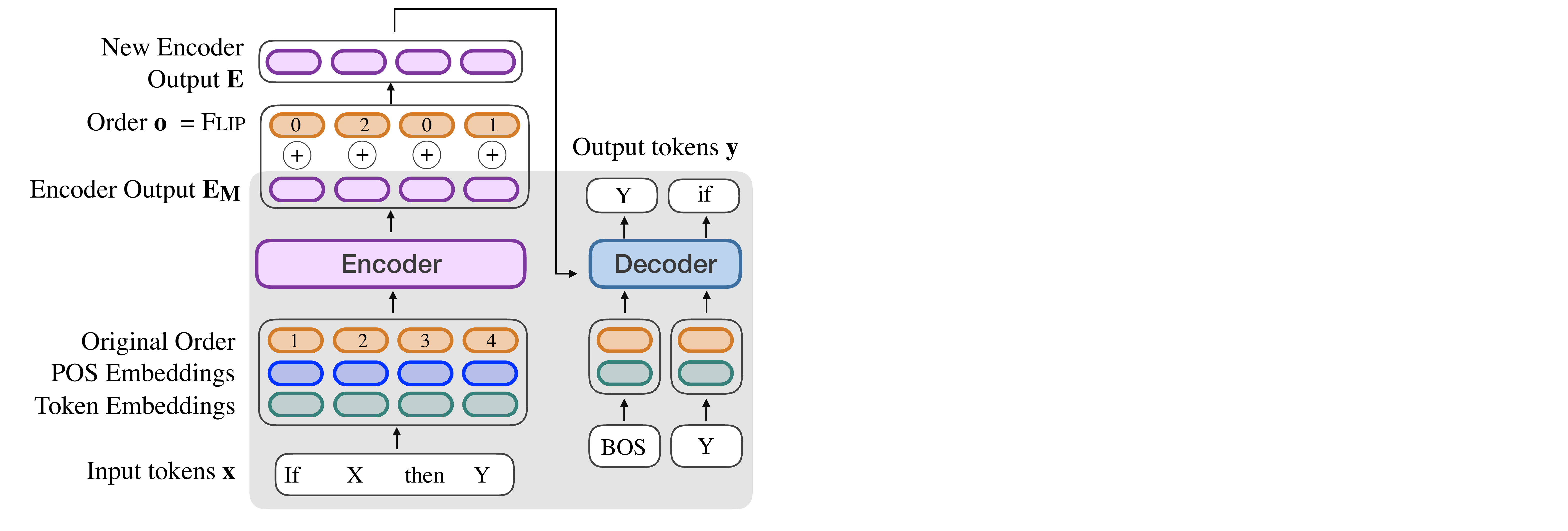}
    \caption{Source Order reWriting (\textsc{Sow}) model. Our model encodes order preference \textsc{Monotone} or \textsc{Flip} through position embeddings added to the encoder output.} 
    \label{fig:sow-model}
\end{figure}

\section{\textsc{Sow} Model Architecture}
Figure \ref{fig:sow-model} provides an overview of the \textsc{Sow} seq2seq model. We add POS tag embeddings (or corresponding constituent label embeddings for abstracted X and Y) to the input token embeddings and original order position embeddings. As outlined in Section \ref{sec:rearrangement-model}, another set of position embeddings corresponding to the order preference, either \textsc{Monotone} or \textsc{Flip}, are further added to the output of the final layer of the encoder. The decoder attends over these augmented encodings during both training and inference.

\section{Syntactic Exemplars}
\label{sec:syntactic_exemplars}
Table \ref{table:mingda-ex} provides an example from the test set of \citet{chen-etal-2019-controllable}. The output retains the semantics of the input sentence while following the structure of the exemplar.
\begin{table}[H]
\small
\begin{tabular}{l} \toprule
    I: his teammates’ eyes got an ugly, hostile expression. \\
    E: the smell of flowers was thick and sweet.\\
    O: the eyes of his teammates had turned ugly and hostile.\\
    \bottomrule
\end{tabular} 
\caption{Example of input (I), syntactic exemplar (E), and the reference output (O) from the evaluation test set of \cite{chen-etal-2019-controllable}.}
\label{table:mingda-ex} 
\end{table}

\section{Example Generations}
In Table \ref{table:example-generations}, we provide examples of paraphrases generated by our system (\textsc{Sow-Reap}) and the baseline SCPN \cite{iyyer2018adversarial} system. We additionally include the phrase level transductions applied to obtain the sentence level reordering by our system (column 1) and the input template that the corresponding SCPN generation was conditioned on (Column 3).

\begin{table*}
\small
\begin{tabular}{p{4.4cm}|p{3.5cm}|p{3cm}|p{3.35cm}}
\toprule
Rules (\textsc{Sow}) & Output (\textsc{Reap}) & Template (SCPN) &  Output (SCPN)\\ \midrule
\multicolumn{4}{l}{\textbf{Input}: the public tender result message normally contains the following information :} \\ \midrule
NP normally contains the following NN: $\rightarrow$ the following NN usually contains in NP : & the following information shall normally be included in the public procurement report : & SBARQ ( WHADVP SQ  . )  & where is the public procurement report report usually contains the following information .\\
& & & \\
NP normally VP : $\rightarrow$ usually VP , NP \newline
VBZ the following NN $\rightarrow$ the NN VBZ  & normally the following information shall be included in the public procurement result report :  & S ( PP  , NP  VP . ) & in the public competition , the report on competition contains the following information . \\ \midrule

\multicolumn{4}{l}{\textbf{Input}: the story of obi-wan kenobi ends here . } \\ \midrule 
NP VP . $\rightarrow$ VP is NP \newline the NN of NP $\rightarrow$ NP NN . & end of the obi-wan kenobi story . & S ( VP . )  & tell the story of obi-wan kenobi . \\ & & & \\
the story PP NNS here . $\rightarrow$ there NNS a story PP . & here ends the story of obi-wan kenobi . &  S ( S , CC  S . )  & the story of obi-wan kenobi is here , and it ends here . \\ \midrule

\multicolumn{4}{l}{\textbf{Input}: i leased it before i knew where the money came from .  } \\ \midrule
i VBN it before i VP . $\rightarrow$ before i VP , i VBN it .  & before i knew where the money came from , i rented it . &   SBARQ ( WHADVP SQ  . )   & where did you learn that it was the money  ? \\ & & & \\
NP knew SBAR . $\rightarrow$ SBAR , S knew . & where the money came from , i lent it to me before i knew . &  S ( NP  VP . )  & i borrowed money before i knew where the money came from .\\ \midrule

\multicolumn{4}{l}{\textbf{Input}: priority actions should be more clearly specified in future reviews . } \\ \midrule

NP should be more clearly specified PP . $\rightarrow$ PP , NP should be clearly specified . & in future reviews , priority measures should be more clearly specified . &   S ( S , CC S . )   & priority actions should be more clearly specified in future reviews , and they should be informed . \\ & & & \\
ADVP VBN in future reviews $\rightarrow$ VBN in future reviews ADVP  & priority measures should be specified in future reviews clearly . &  SBARQ ( WHADVP SQ  . )  & where should priority actions are more clearly specified in future reviews ? \\ \midrule

\multicolumn{4}{l}{\textbf{Input}: okay , well , tonight the occasion is calling . } \\ \midrule
ADJP , S . $\rightarrow$ S , ADJP . \newline well , NN the occasion VP $\rightarrow$  the occasion VP , NN  & the occasion is calling today , okay ?  &  S ( NP VP . )   & the opportunity is calling . \\ & & & \\
ADJP , S . $\rightarrow$ S , ADJP . \newline well , NP VBZ calling $\rightarrow$ VBZ calling NP & we 'll call it tonight , okay ?  &    S ( ADVP NP VP . )  & of course , the occasion is calling . \\ \midrule


\multicolumn{4}{l}{\textbf{Input}: a minor risk considering the number of telephones in new york . } \\ \midrule
a JJ risk considering NP . $\rightarrow$ NP is a JJ risk . \newline the NN of NP $\rightarrow$ NP NN & phones in new york are a minor risk considering . & SBARQ ( WHADVP SQ  .)  & when do you consider the number of telephones in new york ? \\ & & & \\
NP$_1$ considering NP$_2$ . $\rightarrow$ considering NP$_2$ for NP$_1$ \newline NN of NP $\rightarrow$ NP NN \newline  NP in JJ york $\rightarrow$ JJ york NP  & in new york , the number of phones is a minor risk . &   FRAG ( SBAR ) .  & that minor risk is the number of telephones in new york . \\ \midrule

\multicolumn{4}{l}{\textbf{Input}: that dress gets me into anywhere i want .  } \\ \midrule
that S i VBP . $\rightarrow$ i VBP S .& i want that dress gets me into the place . &  NP ( NP . )  & that dress gets me in there , i wish . \\ & & & \\
that S i VBP . $\rightarrow$ i VBP S . \newline NN gets me PP $\rightarrow$ PP , NN gets me . & i want a dress in front of me . &    S ( VP . )   & i want everywhere . \\
\bottomrule
\end{tabular}
\caption{Examples of paraphrases generated by our system and the baseline SCPN model. The outputs from our model successfully rearranges the different structural components of the input sentence to obtain meaningful rearrangements. SCPN on the other hand tends to conform to pre-specified templates that are often not aligned with a given input.}
\label{table:example-generations}
\end{table*}

\section{Implementation Details}
The hyperparameters values used in \textsc{reap} (see Table \ref{table:params-reap}) and \textsc{Sow} (see Table \ref{table:params-sow}) models. Note that we do not use coverage loss for the \textsc{Sow} model.

\begin{table}[H]
    \small
    \begin{tabular}{l|l}\toprule
        \multicolumn{2}{l}{Seq2seq transformer architecture}  \\ \midrule
        Hidden size & 256 \\
        Num layers & 2 \\
        Num heads & 8 \\
        Dropout & 0.1 \\ \midrule
        \multicolumn{2}{l}{Training} \\
        \midrule
        Optimizer & Adam, $\beta=(0.9, 0.999), \epsilon=10^{-8}$\\
        Learning rate & 0.0001\\
        Batch size & 32 \\
        Epochs & 50 (maximum) \\
        Coverage loss coeff. & 1 (first 10 epochs), 0.5 (10 - 20\\
        &  epochs), 0 (rest)\\ \midrule
        \multicolumn{2}{l}{Inference} \\ \midrule
        $k$ in top-$k$ & 20 \\
        Beam Size & 10 \\
        \bottomrule
    \end{tabular}
    \caption{Hyperparameters used in the implementation of the \textsc{Reap} model. }
    \label{table:params-reap}
\end{table}

\begin{table}[H]
    \small
    \begin{tabular}{l|l}\toprule
        \multicolumn{2}{l}{Seq2seq transformer architecture}  \\ \midrule
        Hidden size & 256 \\
        Num layers & 2 \\
        Num heads & 8 \\
        Dropout & 0.1 \\ \midrule
        \multicolumn{2}{l}{Training} \\
        \midrule
        Optimizer & Adam, $\beta=(0.9, 0.999), \epsilon=10^{-8}$\\
        Learning rate & 0.0001\\
        Batch size & 32 \\
        Epochs & 50 (maximum) \\
         \midrule
        \multicolumn{2}{l}{Recombination of rules/transductions} \\ \midrule
        Ignored tags & DT, IN, CD, MD, TO, PRP \\ 
        Max. no. of rules & 3 \\
        \bottomrule
    \end{tabular}
    \caption{Hyperparameters used in the implementation of the \textsc{Sow} model.}
    \label{table:params-sow}
\end{table}

\end{document}